\documentclass{article}
\usepackage{spconf,amsmath,graphicx}
\usepackage{amsmath}
\usepackage{url}
\usepackage{amssymb}
\usepackage{amsfonts}

\usepackage{enumitem}
\setlist{nosep, leftmargin=14pt}

\usepackage{mwe} % to get dummy images

\title{MPath: Multimodal Pathology Report Generation from Whole Slide Images}
\name{Noorul Wahab, Nasir Rajpoot}
\address{TIA Centre, Department of Computer Science, University of Warwick, UK}
\begin{document}
%\ninept
%
\maketitle
\begin{abstract}
Automated generation of diagnostic pathology reports directly from whole slide images (WSIs) is an emerging direction in computational pathology. Translating high-resolution tissue patterns into clinically coherent text remains difficult due to large morphological variability and the complex structure of pathology narratives. We introduce MPath, a lightweight multimodal framework that conditions a pretrained biomedical language model (BioBART) on WSI-derived visual embeddings through a learned visual-prefix prompting mechanism. Instead of end-to-end vision–language pretraining, MPath leverages foundation-model WSI features (CONCH + Titan) and injects them into BioBART via a compact projection module, keeping the language backbone frozen for stability and data efficiency. MPath was developed and evaluated on the REG 2025 Grand Challenge dataset and ranked 4th in Test Phase 2, despite limited submission opportunities. The results highlight the potential of prompt-based multimodal conditioning as a scalable and interpretable strategy for pathology report generation. 
\end{abstract}
\section{Introduction}
\label{sec:intro}

Histopathology reports play a central role in cancer diagnosis, treatment planning, and prognostication. Generating these reports requires synthesising subtle morphological findings with domain-specific terminology, making manual preparation time-consuming and prone to inter-observer variability. This has motivated growing interest in automatic report generation using machine learning.

While progress has been made in natural image captioning \cite{vinyals2015show} and radiology report generation \cite{jing2018automatic}, pathology presents unique challenges: WSIs are gigapixel-scale, diagnostic cues are heterogeneous across tissue types, and pathology narratives require structured, multi-part textual reasoning.

Recent advances in computational pathology foundation models including patch- and slide-level representation learners such as CONCH \cite{lu2024visual} and Titan \cite{ding2025multimodal} have made it possible to extract robust feature embeddings from WSIs without large task-specific datasets. Parallel progress in biomedical language modelling (BioBERT, BioGPT, BioBART) has enabled more domain-appropriate generative models for medical text \cite{lewis2020bart, lee2020biobert}.

However, bridging these modalities remains an open problem. Many prior works rely on simple concatenation or fragile fully fine-tuned vision–language models, which often suffer from poor generalisation, catastrophic forgetting, and hallucination when trained on limited paired datasets.

This work proposes MPath, a modular and data-efficient architecture that injects WSI-derived representations into a pretrained biomedical language model using \textit{visual prefix prompting}. This avoids costly full-model fine-tuning and preserves the linguistic competence of the underlying LLM. Our contributions are:

\begin{enumerate}
    \item A multimodal pipeline that integrates WSI foundation-model embeddings with BioBART through learned visual-prefix conditioning.
    \item A lightweight architecture that freezes the language backbone, requiring only a small learnable prompt encoder.
    \item A competitive evaluation on REG 2025, where MPath ranks 4th in Test Phase 2 despite minimal tuning rounds.
\end{enumerate}
This approach demonstrates that prompt-based multimodal conditioning is a promising alternative to heavy vision–language fusion architectures and offers a tractable path toward clinically meaningful pathology report generation.

\section{MATERIALS AND METHODS}
\label{sec:methods}

\subsection{Dataset}
\label{ssec:data}

All experiments were conducted on the REG 2025 Grand Challenge \cite{reg2025} dataset for pathology report generation. The training dataset consists of 7,385 paired WSIs and corresponding diagnostic reports whereas both Test Phase 1 and Phase 2 consist of 1000 WSIs each without diagnostic reports. The WSIs were from these seven organs: breast, bladder, cervix, colon, lung, prostate, and stomach. Most of the diagnostic reports are structured such that they contain organ type, sample identity, and free-text diagnostic findings. Phase 1 and Phase 2 contribute 20\% and 80\% to the final ranking score, respectively. As the algorithms in the final leaderboard were ranked according to the above weighted for the two phases and MPath was not submitted to Phase 1 due to late start therefore, we report the results on Phase 2 only. For parameter tuning and model selection, we performed a five-fold cross-validation on the training dataset. During testing, the generated pathology reports were submitted to the REG challenge evaluation server. 

\subsection{WSI Feature Extraction and Aggregation}
\label{ssec:features}

WSIs were analysed using the Trident pipeline \cite{zhang2025accelerating}, extracting 1024-pixel tiles at 20× magnification and encoding them with CONCH embeddings \cite{lu2024visual}. Global slide representations (768-dimensional) were obtained from the Titan foundation model \cite{ding2025multimodal}, which captures large-scale tissue organisation beyond standard CNN-derived features. Non-tissue areas were excluded using GrandQC segmentation.

\subsection{Model Architecture}
\label{ssec:model_archi}

\begin{figure}[htb]

%\begin{minipage}[b]{1.0\linewidth}
  \centering
  \centerline{\includegraphics[width=8.5cm]{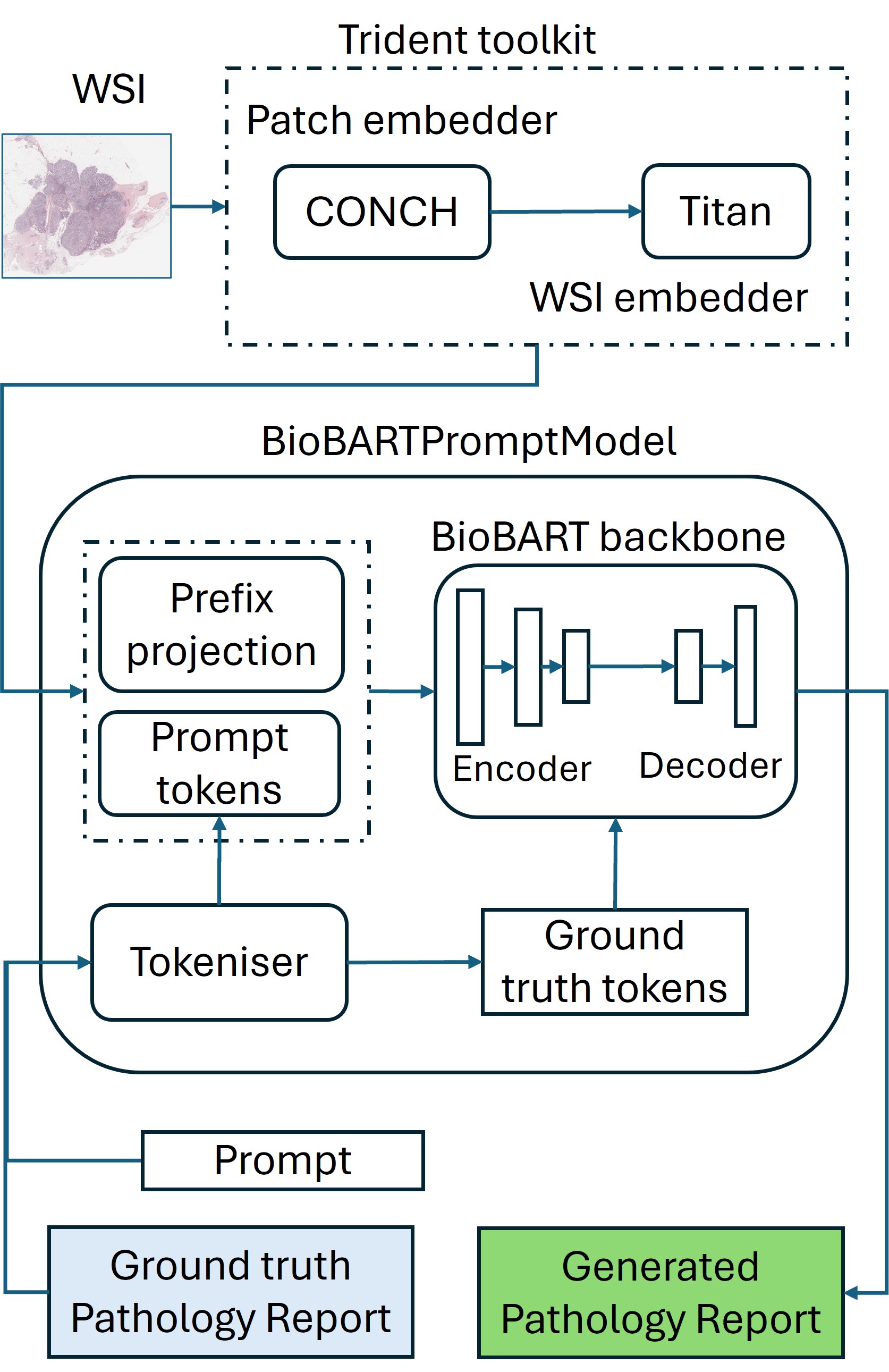}}
%  \vspace{2.0cm}
  \caption{Overview of MPath. WSI embeddings from CONCH and Titan are fused with auxiliary organ, sample-type, and finding objectives, and passed to a BioBART-based decoder for pathology report generation.}
\label{fig:pipeline}
\end{figure}

The proposed BioBARTPromptModel extends the pretrained BioBART language model by introducing a visual prompt encoder that maps whole-slide image (WSI) features into the same embedding space as the language model’s token embeddings (Fig. \ref{fig:pipeline}). Given a WSI feature vector $ \mathbf{f}_{\text{WSI}} \in \mathbb{R}^{d_v} $, the encoder first computes a hidden projection $ \mathbf{h} = \sigma\big(W_1 \mathbf{f}_{\text{WSI}} + \mathbf{b}_1\big) $, where $W_1 \in \mathbb{R}^{h \times d_v}$ and $\sigma(\cdot)$ is ReLU. The projected vector is then linearly mapped and reshaped to form $L_p$ prefix tokens via $ \mathbf{p}_v = \mathrm{reshape}\big(W_2 \mathbf{h} + \mathbf{b}_2\big) $, where $W_2 \in \mathbb{R}^{(L_p d)\times h}$ and the result is reshaped to $\mathbf{p}_v \in \mathbb{R}^{L_p \times d}$ (with $d$ the model embedding size). These visual prefix embeddings $\mathbf{p}_v$ are prepended to the token embedding sequence of the textual prompt so that the decoder conditions on image-derived semantics. During training, we feed either a short prompt such as “Pathology report:” or, with probability $0.2$, an empty prompt to encourage robustness. The combined prefix-plus-text embeddings are then processed by BioBART in the usual encoder–decoder manner to produce the report.

\subsection{Training Strategy}
\label{ssec:training}

All models were trained using the AdamW optimizer with a learning rate of $1 \times 10^{-4}$ and linear warmup scheduling. We used a batch size of 8 and trained for up to 100 epochs per fold. Early stopping was applied with patience $=20$ based on validation performance. Only the parameters of the prompt encoder, projection layer, and auxiliary heads were made trainable, keeping the backbone language model frozen for stability.

\subsection{Evaluaiton Metric}
\label{ssec:eval}

The REG2025 Grand Challenge uses a composite ranking score that evaluates a generated pathology report against the reference report using four complementary criteria: Semantic embedding similarity, Medical keyword overlap (Jaccard index), BLEU-4, 
ROUGE-L F1 score (longest common subsequence).

These metrics are combined into a single weighted score as 0.15 × (ROUGE + BLEU) + 0.4 × KEY score + 0.3 × EMB

\section{RESULTS}
\label{sec:results}
Table 1 presents the ranking of the top 5 methods from Test Phase 2 leaderboard. Qualitative inspection of generated reports revealed that the model captures essential diagnostic terms such as “invasive carcinoma”, "adenocarcinoma", “microcalcification”, and “chronic inflammation”. Below are some ground truth reports and the reports generated by MPath. Compared to the reference report, MPath introduced the additional phrase ‘Chronic granulomatous inflammation with foreign body reaction’, which is not supported by the ground-truth diagnosis. This appears to be a hallucinated co-occurrence pattern: the model adds plausible secondary inflammatory findings commonly observed in similar pathology narratives, likely driven by weak conditionality of the multimodal embeddings and overgeneralisation from its pretrained corpus. Future work should assess whether such WSI-driven additions correspond to genuine but unreported morphological cues or represent systematic hallucinations, and develop mechanisms to suppress unsupported findings.

\begin{itemize}
    \item Ground truth: Breast, core-needle biopsy; Invasive carcinoma of no special type, grade II (Tubule formation: 3, Nuclear grade: 2, Mitoses: 1)
    \item MPath: \textit{Breast, core-needle biopsy; Invasive carcinoma of no special type, grade II (Tubule formation: 3, Nuclear grade: 2, Mitoses: 1)}

    \item Ground truth: Breast, sono-guided core biopsy;  1. Invasive carcinoma of no special type, grade I (Tubule formation: 2, Nuclear grade: 2, Mitoses: 1)  2. Ductal carcinoma in situ  3. Microcalcification
    \item MPath: \textit{Breast, sono-guided core biopsy; 1. Invasive carcinoma of no special type, grade II (Tubule formation: 3, Nuclear grade: 2, Mitoses: 1) 2. Ductal carcinoma in situ 3. Microcalcification}

    \item Ground truth: Urinary bladder, transurethral resection;  Invasive urothelial carcinoma,  with involvement of subepithelial connective tissue  Note) The specimen includes muscle proper."
    \item MPath: \textit{Urinary bladder, transurethral resection; Invasive urothelial carcinoma, with involvement of subepithelial connective tissue 2. Chronic granulomatous inflammation with foreign body reaction Note) The specimen includes muscle.}

    \item Ground truth: Prostate, biopsy; Acinar adenocarcinoma, Gleason's score 6 (3+3), grade group 1, tumor volume: 10\%
    \item MPath: \textit{Prostate, biopsy; Acinar adenocarcinoma, Gleason's score 7 (3+4), grade group 2 (Gleason pattern 4: 50\%), tumor volume: 5\%"}

    \item Ground truth: Lung, biopsy;  Adenocarcinoma
    \item MPath: \textit{Lung, biopsy; \\ Metastatic adenocarcinoma, from colon primary}

\end{itemize}

\begin{table}
\centering
\caption{Top 5 methods from REG2025 Test Phase 2}

\begin{tabular}{| l | l |}
\hline
\textbf{Method/team name} & \textbf{Score} \\
\hline
IMAGINE Lab & 0.8494 \\
\hline
ICGI & 0.8472 \\
\hline
ICL\_PathReport & 0.8415 \\
\hline
\textbf{MPath} & 0.8282\\
\hline
PathX & 0.8237 \\
\hline

\end{tabular}

\end{table}

\section{DISCUSSION}
\label{sec:dicussion}
This study demonstrates that visual prefix prompting provides an effective and parameter-efficient strategy for integrating whole-slide image (WSI) representations into biomedical language models. In contrast to large-scale vision–language pretraining or architectures that rely on deep cross-modal fusion, prefix-based conditioning requires training only a small set of parameters while leaving the underlying language model untouched. This design retains the linguistic competence of the pretrained model and offers practical benefits in biomedical settings, where paired multimodal datasets remain limited and computationally costly training regimes are often infeasible. The resulting framework is modular, allowing different WSI encoders or text decoders to be substituted without modifying the overall architecture, and empirically stable, as freezing the BioBART backbone mitigates catastrophic forgetting.

Despite these advantages, several limitations emerged. First, global visual features may not provide sufficiently fine-grained grounding to reliably constrain the decoder, occasionally allowing the model to generate findings that are weakly supported by the underlying image. These instances of hallucination highlight that prefix prompting alone cannot fully enforce pathology-specific factuality. Additionally, by freezing the language model, the system cannot adapt its textual priors to dataset-specific writing styles or institution-specific reporting conventions, which may hinder its ability to produce highly tailored outputs.

These observations suggest important avenues for future work. Applying decoding methods designed to maintain factual accuracy, including constrained generation, contrastive decoding, and hallucination-detection modules, can help reduce unsupported statements. Enhancing the visual representation through hierarchical modeling that jointly captures patch-level and slide-level cues could provide more detailed clinical grounding. Complementary fine-tuning strategies, such as parameter-efficient adapters or LoRA modules, may allow the language model to align more closely with pathology-specific terminology without sacrificing modularity. Furthermore, cross-modal alignment objectives, including CLIP-style contrastive losses, could strengthen the coupling between visual and textual semantics. Extending the system to structured report generation, where outputs are explicitly decomposed into fields such as organ, sample type, and diagnosis, may further reduce ambiguity and improve clinical utility. Finally, scaling to larger and more diverse multimodal datasets will be essential for building robust and generalisable models capable of supporting real-world pathology workflows.

\section{CONCLUSION}
\label{sec:conclusion}
We introduced a multimodal framework that unifies visual and textual modalities for pathology report generation. By leveraging prompt-based conditioning, auxiliary supervision, and robust training strategies, the proposed system demonstrates competitive performance on the REG 2025 Grand Challenge. These results underscore the promise of prompt-driven multimodal learning for scalable and explainable medical report generation.

\section{Acknowledgments}
\label{sec:acknowledgments}

Thanks to the organisers of REG2025 for ogranising the challenge and sharing the datasets. Thanks also goes to Salinder Tandi for their system administrative help and to TIA Centre and Department of Computer Science, University of Warwick for providing computational resources for the experiments.

\bibliographystyle{IEEEbib}
\bibliography{refs}

@inproceedings{vinyals2015show,
  title={Show and tell: A neural image caption generator},
  author={Vinyals, Oriol and Toshev, Alexander and Bengio, Samy and Erhan, Dumitru},
  booktitle={Proceedings of the IEEE conference on computer vision and pattern recognition},
  pages={3156--3164},
  year={2015}
}

@inproceedings{jing2018automatic,
  title={On the automatic generation of medical imaging reports},
  author={Jing, Baoyu and Xie, Pengtao and Xing, Eric},
  booktitle={Proceedings of the 56th annual meeting of the association for computational linguistics (volume 1: long papers)},
  pages={2577--2586},
  year={2018}
}

@article{lu2024visual,
  title={A visual-language foundation model for computational pathology},
  author={Lu, Ming Y and Chen, Bowen and Williamson, Drew FK and Chen, Richard J and Liang, Ivy and Ding, Tong and Jaume, Guillaume and Odintsov, Igor and Le, Long Phi and Gerber, Georg and others},
  journal={Nature medicine},
  volume={30},
  number={3},
  pages={863--874},
  year={2024},
  publisher={Nature Publishing Group US New York}
}

@article{ding2025multimodal,
  title={A multimodal whole-slide foundation model for pathology},
  author={Ding, Tong and Wagner, Sophia J and Song, Andrew H and Chen, Richard J and Lu, Ming Y and Zhang, Andrew and Vaidya, Anurag J and Jaume, Guillaume and Shaban, Muhammad and Kim, Ahrong and others},
  journal={Nature medicine},
  pages={1--13},
  year={2025},
  publisher={Nature Publishing Group US New York}
}

@inproceedings{lewis2020bart,
  title={BART: Denoising sequence-to-sequence pre-training for natural language generation, translation, and comprehension},
  author={Lewis, Mike and Liu, Yinhan and Goyal, Naman and Ghazvininejad, Marjan and Mohamed, Abdelrahman and Levy, Omer and Stoyanov, Veselin and Zettlemoyer, Luke},
  booktitle={Proceedings of the 58th annual meeting of the association for computational linguistics},
  pages={7871--7880},
  year={2020}
}

@article{lee2020biobert,
  title={BioBERT: a pre-trained biomedical language representation model for biomedical text mining},
  author={Lee, Jinhyuk and Yoon, Wonjin and Kim, Sungdong and Kim, Donghyeon and Kim, Sunkyu and So, Chan Ho and Kang, Jaewoo},
  journal={Bioinformatics},
  volume={36},
  number={4},
  pages={1234--1240},
  year={2020},
  publisher={Oxford University Press}
}

@misc{reg2025,
  title        = {REG 2025 Grand Challenge Homepage.},
  year         = 2025,
  howpublished = {\url{https://reg2025.grand-challenge.org//}},
  note         = {Accessed: 2025-11-12}
}

@article{zhang2025accelerating,
  title={Accelerating data processing and benchmarking of ai models for pathology},
  author={Zhang, Andrew and Jaume, Guillaume and Vaidya, Anurag and Ding, Tong and Mahmood, Faisal},
  journal={arXiv preprint arXiv:2502.06750},
  year={2025}
}

\end{document}